\documentclass[runningheads]{llncs}

\usepackage{balance}
\usepackage{amsfonts} 
\usepackage{bbm}
\usepackage{amsmath}
\usepackage{wrapfig}
\usepackage{multirow}
\usepackage{multicol}
\usepackage{rotating}
\usepackage{booktabs}
\usepackage{tabularx}
\usepackage{colortbl}
\usepackage{graphicx}
\usepackage{comment}

\definecolor{goodColor}{rgb}{0.7,1,0.7}
\definecolor{badColor}{rgb}{1,0.7,0.7}
\definecolor{neutralColor}{rgb}{1,1,0.7}

\usepackage[T1]{fontenc}
\usepackage{graphicx}

\title{MaskPure: Improving Defense Against Text Adversaries with Stochastic Purification}

\author{Harrison Gietz\inst{1}\thanks{Corresponding author: \email{harrygietz@gmail.com}}\orcidID{0009-0001-0742-9219} \and
Jugal Kalita\inst{2}\orcidID{0000-0002-8765-7018}}
\authorrunning{H. Gietz and J. Kalita}
\institute{Louisiana State University, Baton Rouge LA 70803, USA \and
University of Colorado Colorado Springs, Colorado Springs CO 80918, USA}

\begin{document}

\maketitle

\begin{abstract}
The improvement of language model robustness, including successful defense against adversarial attacks, remains an open problem. In computer vision settings, the stochastic noising and de-noising process provided by diffusion models has proven useful for purifying input images, thus improving model robustness against adversarial attacks. Similarly, some initial work has explored the use of random noising and de-noising to mitigate adversarial attacks in an NLP setting, but improving the quality and efficiency of these methods is necessary for them to remain competitive. We extend upon methods of input text purification that are inspired by diffusion processes, which randomly mask and refill portions of the input text before classification. Our novel method, MaskPure, exceeds or matches robustness compared to other contemporary defenses, while also requiring no adversarial classifier training and without assuming knowledge of the attack type. In addition, we show that MaskPure is provably certifiably robust. To our knowledge, MaskPure is the first stochastic-purification method with demonstrated success against both character-level and word-level attacks, indicating the generalizable and promising nature of stochastic denoising defenses. In summary: the MaskPure algorithm bridges literature on the current strongest certifiable and empirical adversarial defense methods, showing that both theoretical and practical robustness can be obtained together. Code is available on GitHub at \url{https://github.com/hubarruby/MaskPure}.
\keywords{Adversarial Robustness\and Certified Robustness\and Deep Learning}
\end{abstract}

\section{Introduction}

Adversarial attacks have been studied in natural language processing for many years. With the increased use of large language models (LLMs) in real-world applications, it has become increasingly important to prevent adversarial inputs from causing incorrect or harmful outputs in these models; small changes in an input can lead to dramatic failures, such as misclassification, hallucination, and generally erroneous output, depending on the model and task.

Diffusion models have recently found great success in computer vision \cite{croitoru_diffusion_2023}, and as a result, interest has grown in applying diffusion models to NLP tasks as well \cite{zou_diffusion_2023}. The intuition behind the generative portion of a diffusion model is that of ``denoising'' or purifying data. Because of this, in computer vision, these models have successfully been used to mitigate adversarial attacks, by adding and subsequently removing partial noise from an input \cite{carlini_certified_2022,nie_diffusion_2022,xiao_densepure_2023}. 

Some limited initial work has explored employing diffusion-inspired defenses to mitigate adversarial attacks in the context of text \cite{li_text_2023-1}, and previous studies have shown that incorporating randomness and stochastic purification has found success in improving robustness \cite{swenor_using_2021,zeng_certified_2021}. Our purification method, MaskPure, is inspired by Li et al. \cite{li_text_2023-1} for improving robustness of text classification. Their approach randomly masks and refills tokens within multiple copies of an input text using BERT \cite{devlin_bert_2019}, followed by using a voting function to determine the final classification output. The idea behind this approach is that the masking and de-masking of tokens mimics the noising and de-noising that occurs in diffusion-based adversarial defenses in the image setting. MaskPure further explores and improves upon this stochastic purification of text by incorporating the use of different voting methods and fine-tuning unique models for mask-filling, rather than using one model for all parts of the purification and classification task. 

We demonstrate the success of our method by testing BERT \cite{devlin_bert_2019} on various adversarial attacks at the character and word levels, and comparing against recent work that leverages random perturbation-based defense \cite{zeng_certified_2021,li_text_2023-1}. We find that MaskPure outcompetes previous methods when employing different voting-based recovery methods, and that it obtains these gains without any adversarial fine-tuning or any knowledge of attacker vocabulary. This is in contrast to works such as Ye et al. \cite{ye_safer_2020}, which rely on knowledge of the attacks being performed in order to employ their defense. When defending against particularly-difficult modern attacks \cite{jin_is_2020,gao_black-box_2018}, our method obtains accuracy-under attack scores as much as $25\%$ higher than previous work (Table \ref{table:agnews_robustness}).

In addition, we leverage past results from Zeng et al. \cite{zeng_certified_2021} to make certifiable guarantees on MaskPure's performance against adversaries. Our study serves as strong evidence in favor of the continued harnessing of stochastic purification methods to improve robustness, based on both positive theoretical and empirical results.

Overall, the contributions of this paper are the following:
\begin{itemize}
    \item We introduce MaskPure, a novel diffusion-inspired defense mechanism, which enhances robustness against adversarial attacks in text classification by utilizing unique voting methods and a novel mask-filling approach, \textit{without} requiring adversarial training.
    \item MaskPure is the first case of diffusion-inspired text purification we are aware of that is used to successfully defend against character-level attacks. At the same time, MaskPure outperforms previous works on empirical robustness when defending against word-level attacks.
    \item We demonstrate that MaskPure is provably certifiably robust, in addition to its strong empirical performance.
\end{itemize}

The rest of this paper starts by discussing related works, followed by establishing the problem of adversarial robustness in a formal manner and presenting the essence of the MaskPure algorithm. The paper continues with details of the experiments performed and results obtained on the AG News and IMDB datasets; first, empirical results are presented, followed by a proof of certified robustness and presentation of results from the experiments used to obtain robustness certificates.

\section{Related Work}

\subsection{Adversarial Attacks in NLP}
Recent surveys on adversarial robustness in NLP \cite{alshemali_improving_2020,goyal_survey_2023} describe many ways adversaries can be generated in NLP settings: for example by changing the input text at the character level (swapping, replacing), at the word level (insertion, deletion, swapping, substitution), and at the sentence level (deleting, injecting, paraphrasing). Among these, some of the most common and effective attacks include those that use a greedy search algorithm, such as Bert-Attack \cite{li_bert-attack_2020}, TextFooler \cite{jin_is_2020}, DeepWordBug \cite{gao_black-box_2018}, and TextBugger \cite{li_textbugger_2019}. Many of these attack styles are readily implemented in various forms in the TextAttack library \cite{morris_textattack_2020}, which we use to measure the robustness of our method when performing text classification. We test our defense method on one character-level attack, Deep Word Bug \cite{gao_black-box_2018}, and two settings of the word-level attack TextFooler \cite{jin_is_2020}.

\subsection{Adversarial defense in NLP}

Many defense methods have been proposed to enhance the robustness of NLP models, including adversarial training \cite{yoo_towards_2021,madry_towards_2019,kurakin_adversarial_2017,miyato_adversarial_2021,zhu_freelb_2020,jiang_smart_2020}, changes to model architecture \cite{alshemali_improving_2020,goyal_survey_2023,sakaguchi_robsut_2017,jones_robust_2020}, and add-ons such as spell-checking \cite{belinkov_synthetic_2018}. While adversarial training has shown effectiveness in enhancing robustness, it also presents challenges: it requires careful creation of adversarial examples and can significantly increase computational resources and training time. The presented defense method MaskPure avoids these potential issues, while also attaining stronger results than recent defenses that do utilize adversarial training (see Section 5 for comparison).

Recent approaches to defending against adversarial texts by incorporating randomness have shown promising results: Swenor and Kalita \cite{swenor_using_2021} demonstrated that adding random perturbations to adversarial inputs can bring classification model performance back to its original level. Additionally, Li et al. \cite{li_text_2023-1} demonstrate one of the first uses of diffusion-inspired purification in the text domain, by masking (``noising'') and replacing (``de-noising'') random tokens in the input. Both of these works show the potential utility of further exploring the use of randomness and purification for mitigating adversarial attacks in language settings. 

Table \ref{tab:defense_comparison} presents a comparison of the MaskPure defense method with other defenses in the literature. In the table, green denotes a positive aspect of the defense method compared to the others, red denotes a negative, and yellow denotes that the aspect of the defense method is neither explicitly positive nor negative compared to the others. ``Defensive logit synthesis'' refers to use of constructed logit scores (e.g. by majority voting or similar methods) rather than only defending with an averaged-logit voting approach. ``Attack levels'' refer to whether the method was tested and/or successful against both word and character-level attacks. ``Fine-tuning requirements'' denote whether or not the method requires fine-tuning of any models apart from the initial fine-tuning of the classifier for the tested dataset. With MaskPure, the masked LM is tuned depending on the task -- this is denoted by ``Per Dataset''; the method used by Li et al. \cite{li_text_2023-1} tunes the same model for masked language modelling and classifying, which requires the least fine-tuning among the listed approaches; for RanMASK, a new classifier must be fine-tuned depending on the relative quantity of the input text that is masked, as well as for each dataset; for SAFER, no fine-tuning is used apart from the classifier tuning.

\begin{table}
    \caption{Comparison of MaskPure with other well-performing defenses. *SAFER's performance, although very strong, benefits from the usage of a synonym table (assuming knowledge of the attack type), making direct comparison unfair.}
    \centering
    \begin{tabularx}{\textwidth}{|>{\hsize=\hsize\centering\arraybackslash}X|>{\hsize=\hsize\centering\arraybackslash}X|>{\hsize=\hsize\centering\arraybackslash}X|>{\hsize=\hsize\centering\arraybackslash}X|>{\hsize=\hsize\centering\arraybackslash}X|}
    \hline
    \textbf{Property} & \textbf{MaskPure (ours)} & \textbf{Text Purification \cite{li_text_2023-1}} & \textbf{RanMASK \cite{zeng_certified_2021}} & \textbf{SAFER \cite{ye_safer_2020}} \tabularnewline
    \hline
    Certifiably Robust & \cellcolor{goodColor}Yes & \cellcolor{badColor}No & \cellcolor{goodColor}Yes & \cellcolor{goodColor}Yes \tabularnewline
    \hline
    Uses Defensive Logit Synthesis & \cellcolor{goodColor}Yes & \cellcolor{badColor}No & \cellcolor{goodColor}Yes & \cellcolor{goodColor}Yes \tabularnewline
    \hline
    Adversarial Training & \cellcolor{goodColor}Not Needed & \cellcolor{badColor}Yes & \cellcolor{badColor}Yes & \cellcolor{badColor}Yes \tabularnewline
    \hline
    Attack Levels & \cellcolor{goodColor}Word \& Char. & \cellcolor{badColor}Word only & \cellcolor{goodColor}Word \& Char. & \cellcolor{badColor}Word only \tabularnewline
    \hline
    Empirical Performance & \cellcolor{goodColor}Very Strong & \cellcolor{neutralColor}Strong & \cellcolor{neutralColor}Strong & \cellcolor{neutralColor}Very Strong* \tabularnewline
    \hline
    Uses Synonym Table & \cellcolor{goodColor}No & \cellcolor{goodColor}No & \cellcolor{goodColor}No & \cellcolor{badColor}Yes \tabularnewline
    \hline
    Fine-tuning Requirements & \cellcolor{neutralColor}MLM, Per Dataset & \cellcolor{goodColor}Classifier and MLM, Together & \cellcolor{badColor}Classifier, Per Dataset \& Per Mask \% & \cellcolor{goodColor}Classifier Only \tabularnewline
    \hline
    \end{tabularx}
    \label{tab:defense_comparison}
\end{table}


\section{Problem Formulation and Algorithm Design}

\begin{figure*}
    \centering
    \includegraphics[width=0.9\linewidth]{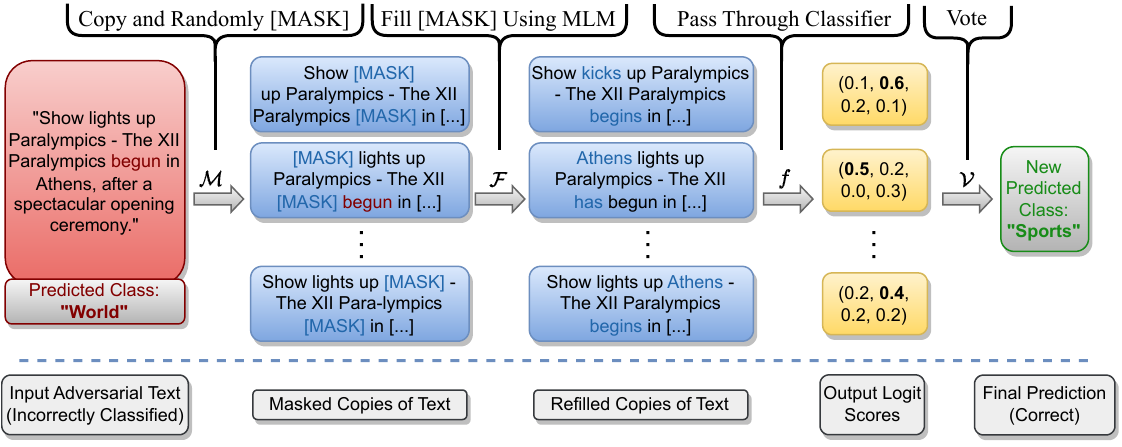}
    \caption{The pipeline for the MaskPure purification process, as demonstrated using an example from the AG News dataset. In this illustrative example, the perturbed sample contains an adversarial word, ``begun'' that leads to misclassification. The masking, filling, and voting process allows the classifier to correctly recover the correct label, Sports.}
    \label{fig:maskpureprocess}
\end{figure*}

Stochastic purification has a history of success in the continuous domain of computer vision \cite{carlini_certified_2022,nie_diffusion_2022,xiao_densepure_2023}, and preliminary promising results for defending against textual adversaries. Our study presents a novel method that improves stochastic text purification methods used to mitigate adversarial attacks. A formal description of the problem and our approach is provided below.

\subsection{Notation for Adversarial Examples}
We use notation similar to Zeng et al. \cite{zeng_certified_2021} and Levine and Feizi \cite{levine_robustness_2020}: a dataset of texts, $\mathcal{X}$, has corresponding class labels, $y\in\mathcal{Y}$. Each $y\in \mathcal{Y}$ is an integer label from the set $C:=\{1, 2, ..., c\}$, where c is the number of classes that can be predicted. Each $x\in \mathcal{X}$ is a sequence of ``tokens'' (typically words, but also including other characters, like punctuation) which can be passed into a trained classifier model, $f: \mathcal{X}\rightarrow\mathcal{Y}$ . Hence, any text $x\in \mathcal{X}$ can be expressed as $x_1, x_2, ..., x_j$, where $j$ is the number of individual tokens in the text.

To formulate the concept of an adversarial input, we consider what results from ``perturbing'' or changing $d$ number of tokens within some $x$. If $x$ is an input that can be correctly classified by $f$ (i.e. $f(x) = y$, where y is the correct class), then a successful adversarial version of the input, called $x^\prime$, is a sequence that differs from $x$ by $d$ tokens while also satisfying $f(x^\prime)\neq y$. In other words, $x^\prime$ is a maligned version of $x$ designed to fool the classifier $f$. 
We use Hamming distance $||\cdot||_0$ to denote the similarity of two input texts; saying that $||x-x^\prime||_0 = d$ is equivalent to saying that $x$ and $x^\prime$ have different tokens at $d$ positions (while being the same in every other position). In this case, $x$ and $x^\prime$ are of the same length, $j$.

We say that the model $f$ is certified robust against $d$-sized adversaries on an input $x$ if, with some (preferably high) probability, we know that $f(x^\prime) = y$. This definition applies for any $x^\prime$ satisfying $||x-x^\prime||_0 \le d$, meaning the model is robust against any and all changes to the text sequence $x$, so long as the number of changes is below or equal to the size $d$.

Next, we define the symbol $\ominus$ between two texts $x$ and $x'$, which represents the set of token indices where these texts differ. To illustrate this, consider the text $x$ as ``Quick Red Fox'' and $x'$ as ``Quick Blue Fox'', the set $x \ominus x'$ would be $\{2\}$ because the tokens at the second position in the two texts are different.

Furthermore, consider a set of indices $\mathcal{S}$, denoted as $\{1, \ldots, j\}$. Let $\mathcal{I}(j, k)$ be a set that contains all sets of $k$ unique indices from $\mathcal{S}$. For example, for $\mathcal{S}$ = $\{1, 2, 3, 4\}$ with cardinality $j=4$, $\mathcal{I}(4, 2)$ might include subsets like $\{1, 2\}$, $\{1, 3\}$, and so on.

Lastly, let $\mathcal{U}(j, k)$ represent a uniform distribution over $\mathcal{I}(j, k)$. In other words, if we sample from $\mathcal{U}(j, k)$, we are effectively selecting $k$ out of $j$ indices without replacement, uniformly. As an example, if we draw a sample from $\mathcal{U}(7, 4)$, we might obtain a set like $\{2, 4, 6, 7\}$.

\subsection{Notation for Text Processing }

Our algorithm involves multiple steps prior to input to the classifier; first, we introduce a mask operation, denoted as $\mathcal{M}$, which maps pairs of texts and indices, $\mathcal{X}$ and $\mathcal{I}(j, k)$, to a new set $\mathcal{X}_{\text {mask }}$. $\mathcal{X}_{\text {mask }}$ is a similar set to $\mathcal{X}$, but some words in the texts are replaced by a [MASK] token. In particular, every word whose index does not correspond to a value in the input set of indices is converted to the [MASK] token. To illustrate this, consider the text ``Hello Beautiful World'' and the input indices $\{1,3\}$; in such a case, $\mathcal{M}$ would transform the text to ``Hello [MASK] World''.

Next, we define a new function $\mathcal{F}$, which operates on $\mathcal{X}_{\text {mask }}$ and produces $\mathcal{X}_{\text {fill }}$ by replacing [MASK] tokens with predicted words from a masked language model (MLM). 

Returning to our classifier, we use $f: \mathcal{X}_{\text {fill }} \rightarrow \mathcal{Y}$, as our base method for classifying texts, where the predicted class is represented by $y\in \{1, 2, ..., c\}$. In our case, $f$ is a pre-trained BERT classification model from the textattack library.

To summarize, the pipeline for processing a single text involves the following mappings, in sequential order:
\begin{equation}
\mathcal{X} \times \mathcal{I}(j, k) \xrightarrow[\text{}]{\mathcal{M}} \mathcal{X}_{\text {mask }} \xrightarrow[\text{}]{\mathcal{F}} \mathcal{X}_{\text {fill }} \xrightarrow[\text{}]{f} \mathcal{Y}.
\label{algo_mapping}
\end{equation}


In practice,  $f$ can be considered to be composed of two parts, such that $f = f_c \circ f_l$. The first part of the function, $f_l$ outputs a vector of $c$ logit scores ranging from 0 to 1, which collectively sum to 1. Following that, $f_c$ takes an argmax over the output of $f_l$ to determine the index of the highest logit score, and this returned index is considered the predicted class $y = f(x)$. 

Some of the voting methods employed using MaskPure (which are denoted by $\mathcal{V}$ and discussed later on) skip the final step of $f_c$, and instead use a collection of outputs of $f_l$ to make the decision of the predicted class.

For ease of expressing and proving our claims about certified robustness, we re-frame this notation. Similar to \cite{zeng_certified_2021}, we simplify by defining $p_c(x)$ as the probability that, after randomly masking and filling, $f$ returns the class $c$:
$$
p_c(x)=\underset{\mathcal{H} \sim \mathcal{U}\left(j_{x}, k_{x}\right)}{\mathbb{P}}(f(\mathcal{F}(\mathcal{M}(x, \mathcal{H})))=c) .
$$
In this equation, $j_x$ is the cardinality (number of tokens) of $x$, and $k_x$ is the number of tokens in $x$ to be left unmasked. I.e. $k_x :=r_{int}( (1-m)\cdot j_x)$, with $m$ being the chosen proportion of tokens to be masked and $r_{int}(\cdot)$ indicating a nearest-integer rounding function. In our experiments, we set $m = 0.3$. Details surrounding hyperparameter selection are excluded due to space constraints.

Following similar steps to Zeng et al. \cite{zeng_certified_2021}, we then define a composite ``smoothed'' classifier $g(x)$ as:
$$
g(x)=\underset{c \in \mathcal{Y}}{\arg \max }\left[p_c(x)\right].
$$
Intuitively, $g(x)$ represents the most probable output from $f(x)$, if all but $k_x$ words from $x$ are randomly masked and re-filled before passing through the classifier.

\subsection{Miscellaneous Notation} The following notation is used in the discussion of our algorithm below. Let $I_n = \{1, 2, ..., n\}$. For a set of $n$ ordered text inputs, called $X_n$, and a set of $n$ ordered samples $\mathcal{H}\sim \mathcal{U}(j, k)$ called $H_n$, define $\phi: I_n \rightarrow X_n\times H_n$ by $\phi(i) = (x_i, \mathcal{H}_i)$ for each $i \in I_n$.

\subsection{MaskPure Algorithm}
The purification method aims to be a simple algorithm leveraging an existing approach by Li et al. \cite{li_text_2023-1}. Note that the method is agnostically applied to all inputs, since in real life settings it is difficult to detect which inputs are adversarial or not. The general structure of the purification methods is as follows, for any one input:
\begin{enumerate}
    \item Make $v$ identical copies of the input $x$, such that we have a set of copies, $X = \{x_1, x_2, ..., x_v\}$; for each copy $x_i$, mask a proportion $k_x$ of the input tokens according to the masking scheme, $\mathcal{M}$. That is, take $v$ samples $\mathcal{H}\sim \ U(h_x, k_x)$, such that we have $H = \{\mathcal{H}_1, \mathcal{H}_2, ..., \mathcal{H}_v\}$; then, obtain a set of masked outputs, $\mathcal{M}\big(\phi(I_v)\big)$. The function $\phi$ is defined as above.
    \item Refill the masked tokens in each of the $v$ copies, according to the mask-filling scheme, $\mathcal{F}$. In our case, $\mathcal{F}$ uses a masked language model to predict words to replace each [MASK] token.
    \item Pass the new ``re-filled'' copies of the input text through the classification model $f$ (in the case with averaged logit voting, only pass through $f_l$, instead of the entire composite function $f_c\circ f_l$). Then, use a voting function $\mathcal{V}$ to obtain a final set of output logit scores which conveys the top predicted class.
\end{enumerate}
Hence, for a classification task with $c$ output classes, the final predicted output $c^\prime \in \{1, 2, .., c\}$ can be expressed as 
\begin{equation}
    c^\prime = argmax\Big\{\mathcal{V}\Big(f\Big(\mathcal{F}\Big(\mathcal{M}\big(\phi(I_v)\big)\Big)\Big)\Big)\Big\}.
\end{equation}


\section{Experiments}

\subsection{Datasets}
We measure the model's robustness to adversaries on the AG News sentiment classification task \cite{zhang_character-level_2015} and the IMDB movie review classification task \cite{maas_learning}, since these are widely used tasks for classification models, making results easily comparable to other works. For AG News, we use the initial 1000 test samples for each dataset provided in the TextFooler Github repository \cite{jin_is_2020}, to replicate results from Li et al. \cite{li_text_2023-1}. For the IMDB task, 100 samples were randomly drawn from the test set used by Jin et al. \cite{jin_is_2020}. This smaller sample number is due to the larger amount of text in each sample, which in turn requires more computational power and longer processing time.

\subsection{Models}
Classifications outputs are produced using \textit{bert-base-uncased-ag-news}, available from the TextAttack library \cite{morris_textattack_2020}; this model has been fine-tuned for text-classification on the AG News training corpus.
To perform the mask-filling process, we use \textit{bert-based-uncased} for masked language modelling, provided by Huggingface \cite{wolf_huggingfaces_2020}, and fine-tuned on the training data from the HuggingFace AG News dataset using a cross entropy loss function. 

\subsection{Implementation Details}

One key variation behind our method compared to Li et al. \cite{li_diffusion_2023} comes from fine-tuning the masked LM from step (2) on the dataset being tested on, rather than using the baseline BERT model. The intuition behind this is based on diffusion purification in computer vision; since the goal of an adversarial purification process is to remove noise and faults in a text, it makes sense for the purification process to bring the perturbed sample closer to the original distribution of data. Hence, we should expect the model used for mask filling to contribute to better performance if it is able to better fill masks according to the structure of the original data.

This is different from Li et al, \cite{li_text_2023-1}, where the authors take a combined-training approach; in their method, the same model is used for both classification and mask filling, and it is trained on a joint loss function based on cross entropy; this cross entropy is calculated based on both classification and mask filling performance. We hypothesize that the reason for MaskPure's better performance may be that this combined training actually hinders performance when compared with fine-tuning two separate models on each task; the jointly-trained model is required to optimize for two distinct components of a loss function, which may compete against one another. Notably, using the alternative approach, MaskPure obtains improved performance on AG News and IMDB when compared with \cite{li_text_2023-1}, and it does this \textit{without} including adversarial-training of the classifier or mask-filling algorithm. 


The other factor that differentiates MaskPure involves the voting process, $\mathcal{V}$. It is common practice for adversarial defense methods to use multiple modified copies of an input for classification, obtaining predictions from each copy, and using a voting process to combine the predictions \cite{li_searching_2021,swenor_using_2021,li_diffusion_2023,zeng_certified_2021}. Our result takes advantage of this approach, evaluating accuracy-under attack using different voting methods. These methods include logit averaging, majority vote-based logit scores, and a naive max or ``one hot'' majority vote-based logit scores. A description of each is provide below.

The logit averaging function takes the average of the $n$ logit scores which are outputted by the classifier model. For example, assume there is a case with $n=5$ vectors of logits, $s_1$ through $s_5$, where there are 2 classes scored in each vector. If the vectors are $\mathbf{s}_1=(0.9,0.1)$, $\mathbf{s}_2=(0.76,0.24)$, $\mathbf{s}_3=(1,0)$, $\mathbf{s}_4=(0.81,0.19)$, $\mathbf{s}_5=(0.16,0.84)$, then the averaged logits are $(0.726, 0.274)$, meaning the first class is the predicted output. 

The majority vote-based logit scores are calculated by considering the top prediction of each of the $n$ copies of the text, and then summing the total number of top predictions for each class. The final output is then normalized based on the number of voters. Using the same example logit scores, since there are 4 ``votes'' for the first class and only 1 vote for the second, that means the final logits outputted would be $(\frac{4}{5}, \frac{1}{5}) = (0.8, 0.2)$.

When using the naive max logit voting method, the result is very similar, but the output is not weighted. Instead, the output would simply be $(1, 0)$, with all of the weight put on the first class, since it won the majority vote.

Our results in Table \ref{li2023_table} show that naive max logit scores perform better than basic majority voting, and that both of the majority voting methods perform better than the averaged logit voting. This informs the view that better majority-vote-resistant attacks need to be discovered to keep up with defenses that trick the typical greedy-search algorithms of attack methods (as also suggested by Devvrit et al. \cite{devvrit_voting_2020}). 

\section{Results} 

\begin{table}[ht]
\caption{TAE after-attack accuracy for BERT on the IMDB and AG News datasets, using different defense methods. The performance of other defense methods are presented as reported by Li et al. \cite{li_text_2023-1}.}
\centering
\begin{tabularx}{\textwidth}{@{}l*{6}{>{\centering\arraybackslash}X}@{}}
\toprule
Defense $\downarrow$ \ Attack $\rightarrow$ & \multicolumn{2}{c}{\textbf{DeepWordBug}} & \multicolumn{2}{c}{\textbf{TextFooler (50)}} & \multicolumn{2}{c}{\textbf{TextFooler (12)}} \\
\cmidrule(lr){2-3} \cmidrule(lr){4-5} \cmidrule(lr){6-7}
& Cln\% & Boa\% & Cln\% & Boa\% & Cln\% & Boa\% \\
\midrule
\textbf{IMDB} $\downarrow$ \\
BERT (No Defense) & \textbf{92.7} & 33.4 & 92.7 & 0.9 & 92.7 & 5.3 \\
Adv-HotFlip \cite{ebrahimi_hotflip_2018} & - & - & 95.1 & 8.0 & 95.1 & 36.1 \\
FreeLB \cite{zhu_freelb_2020} & - & - & \textbf{96.0} & 7.3 & \textbf{96.0} & 30.2 \\
FreeLB++ \cite{li_searching_2021} & - & - & 93.2 & 45.3 & - & - \\
Text Purification \cite{li_text_2023-1} & - & - & 93.0 & 51.0 & 93.0 & 81.5 \\
MaskPure: Averaged Logit & 92.0 & 63.0 & - & - & 92.0 & 73.0 \\
MaskPure: Majority-V Logit & 91.0 & 71.0 & - & - & 91.0 & 75.0 \\
MaskPure: Naive Max Logit & 92.0 & \textbf{78.0} & 94.0 & \textbf{64.0} & 94.0 & \textbf{82.0} \\
\addlinespace
\midrule
\textbf{AG News} $\downarrow$ \\
BERT (No Defense) & 95.9 & 37.1 & \textbf{95.9} & 16.5 & 95.9 & 29.5 \\
Adv-HotFlip \cite{ebrahimi_hotflip_2018} & - & - & 91.2 & 18.2 & 91.2 & 35.3 \\
FreeLB \cite{zhu_freelb_2020} & - & - & 90.5 & 20.1 & 90.5 & 40.1 \\
SAFER \cite{ye_safer_2020} & 95.2 & 78.4 & - & - & - & - \\
Text Purification \cite{li_text_2023-1} & - & - & 90.6 & 34.9 & 90.6 & 61.5 \\
RanMASK \cite{zeng_certified_2021} & 95.3 & 77.1 & 93.9 & 68.6 & - & - \\
MaskPure: Averaged Logit & \textbf{96.3} & 68.9 & 95.6 & 53.5 & 95.9 & 63.9 \\
MaskPure: Majority-V Logit & 95.9 & 79.2 & 95.6 & 69.3 & \textbf{96.0} & 76.3 \\
MaskPure: Naive Max Logit & 95.7 & \textbf{87.8} & 95.6 & \textbf{84.0} & 95.8 & \textbf{85.8} \\
\bottomrule
\end{tabularx}
\label{li2023_table}
\end{table}

See Table \ref{li2023_table} for results comparing MaskPure with other recent stochastic-based defenses. Following the terminology used by Zeng et al. \cite{zeng_certified_2021}, Cln\% refers to the accuracy of the method used \textit{before} attacking was conducted, while Boa\% denotes ``robust accuracy'' of the model while under attack. Measuring both of these values allows us to see how MaskPure performs at defending against adversarial attacks, while at the same time measuring the extent to which it can retain baseline performance of the BERT model used. 

Informed by Si et al. \cite{si_better_2021}, we use the standard Transfer Adversarial Evaluation, TAE (as opposed to Static Attack Evaluation, SAE) to measure the adversarial performance of our model. This evaluation method makes the defense more challenging, as new adversarial texts are generated for every defense type and variation, rather than using only one original set of adversarial texts for evaluation.

In Table \ref{li2023_table}, spaces marked with ``-'' mean that the test results were not available for that particular attack/defense combination. The value k (denoted with ``TextFooler (k)'') corresponds to the size of the synonym list used by the TextFooler attack. Note that \cite{li_text_2023-1} did not test their diffusion-inspired defense against DeepWordBug attacks, making MaskPure the first study we are aware of to test the capability of stochastic-purification on character-level attacks. Zeng et al. \cite{zeng_certified_2021} did test against DeepWordBug, but their work only involved random masking without the diffusion-inspired mask-filling process. 

The tests for MaskPure on AG News were conducted with the 1000-sample test set used by Li et al. \cite{li_text_2023-1}, originally from Jin et al. \cite{jin_is_2020}. Due to resource and time limitations, the tests for MaskPure on IMDB were conducted with 100 random samples from the IMDB test set. To obtain results for this table using MaskPure, the masking rate $m$ is set to 0.3, and voting quantity $v$ is set to 11. 


Of particular note, MaskPure exceeds the performance of Li et al. \cite{li_text_2023-1} by over 25\% when tested on TextFooler with a synonym list of size k = 12, while also retaining high clean accuracy. For all three attacks tested, the top MaskPure robust accuracy score was over 10\% higher than the compared works. The only exception to this 10\%+ robustness increase is when comparing against SAFER \cite{ye_safer_2020} on the AG News dataset; however, this comparison is not entirely fair, due to SAFER's assumed knowledge of attacker vocabulary (as discussed in an earlier section). Hence, MaskPure's ability to outperform SAFER at all (87.8 vs 78.4 Boa\%), even without access to attacker vocabulary, is a testament to the method's effectiveness.

\section{Certified Robustness of MaskPure}

Recent work in computer vision has been able to demonstrate some theoretical justifications for the success of diffusion purification \cite{xiao_densepure_2023,nie_diffusion_2022}. 
For instance, Nie et al. \cite{nie_diffusion_2022} prove that, under certain constraints, the L2 distance between a diffusion-purified adversarial sample and the original clean sample is bounded (with some probability) by a reasonably small value.
Similar (though not entirely analogous) robustness findings have been shown for textual adversaries \cite{zeng_certified_2021,ye_safer_2020}. The approach used by Zeng et al. \cite{zeng_certified_2021} demonstrates certifiable robustness for a classifier trained on samples of text that are partially masked (but not re-filled). Their work can easily be transposed to a similar result in our case, which also includes the ``refilling'' component of the text purification. Our theorem is a direct corollary of theirs, with the only difference being the definition of $\beta$. We demonstrate this below.

\begin{theorem}
    For an original text $x$ and an adversarial text $x^{\prime}$, if $\left\|x-x^{\prime}\right\|_0 \leq d$, then:
\begin{equation}
p_c(x)-p_c\left(x^{\prime}\right) \leq \beta \Delta
\label{thm1}
\end{equation}
$\forall c \in \mathcal{Y}$. \text{Here,}
\begin{equation}
\begin{gathered}
    \Delta=1-\frac{\left(\begin{array}{c}
j_x-d \\
k_x
\end{array}\right)}{\left(\begin{array}{c}
j_x \\
k_x
\end{array}\right)}, \\
\beta=\mathbb{P}\left(f(\mathcal{F}(\mathcal{M}(x, \mathcal{H})))=c \mid \mathcal{H} \cap\left(x \ominus x^{\prime}\right) \neq \emptyset \right) .
\end{gathered}
\end{equation}
\end{theorem}

\textit{Proof}. 
Recall that
\begin{align}
p_c(x) & =\mathbb{P}(f(\mathcal{F}(\mathcal{M}(x, \mathcal{H})))=c) \text { and } p_c\left(x^{\prime}\right) =\mathbb{P}\left(f\left(\mathcal{F}\left(\mathcal{M}\left(x^{\prime}, \mathcal{H}\right)\right)\right)=c\right) .
\end{align}
Using the law of total probability, we can find that:
\begin{align}
p_c(x) &= \mathbb{P}\Big([f(\mathcal{F}(\mathcal{M}(x, \mathcal{H})))=c] \wedge \left[\mathcal{H} \cap\left(x \ominus x^{\prime}\right)=\emptyset\right]\Big) \notag \\ 
&+ \ 
\mathbb{P}\Big([f(\mathcal{F}(\mathcal{M}(x, \mathcal{H})))=c] \wedge \left[\mathcal{H} \cap\left(x \ominus x^{\prime}\right) \neq \emptyset\right]\Big) , \notag \\
p_c\left(x^{\prime}\right) &= \mathbb{P}\Big([f(\mathcal{F}\left(\mathcal{M}\left(x^{\prime}, \mathcal{H}\right)\right))=c] \wedge \left[\mathcal{H} \cap\left(x \ominus x^{\prime}\right)=\emptyset\right]\Big) \notag \\
&+ \ \mathbb{P}\Big([f(\mathcal{F}\left(\mathcal{M}\left(x^{\prime}, \mathcal{H}\right)\right))=c] \wedge \left[\mathcal{H} \cap\left(x \ominus x^{\prime}\right) \neq \emptyset\right]\Big) .
\end{align}

In the case where $\mathcal{H} \cap\left(x \ominus x^{\prime}\right)=\emptyset$ and $\mathcal{F}$ is deterministic (the latter of which we can assume), it is clear that $x$ and $x^{\prime}$ hold the same values at each $ i\in \mathcal{H}$. Hence, conditional on $\mathcal{H} \cap\left(x \ominus x^{\prime}\right)=\emptyset$, it is true that $\mathcal{F}(\mathcal{M}(x, \mathcal{H}))=\mathcal{F}(\mathcal{M}\left(x^{\prime}, \mathcal{H}\right))$. This gives us:
\begin{align}
\mathbb{P}\left(f(\mathcal{F}(\mathcal{M}(x, \mathcal{H})))=c \mid \mathcal{H} \cap\left(x \ominus x^{\prime}\right)=\emptyset\right)&= \notag \\
\mathbb{P}\left(f(\mathcal{F}\left(\mathcal{M}\left(x^{\prime}, \mathcal{H}\right)\right))=c \mid \mathcal{H} \cap\left(x \ominus x^{\prime}\right)=\emptyset\right)&
\end{align}
The rest of the proof follows using the same steps as Zeng et al. \cite{zeng_certified_2021}.

Note that given the black-box nature of language models, it is intractable to precisely compute $p_c(x)$. Instead, we take a similar approach to \cite{jia_certified_2019}, \cite{cohen_certified_2019}, and \cite{zeng_certified_2021}, since we can estimate the guaranteed lower bound of the probability based on the one-sided exact (Clopper Pearson) interval \cite{clopper_use_1934}. More specifically, it is possible to obtain a lower bound on $p_c(x)$ by running the classifier $f$ on $n$ different masked and subsequently-filled copies of an input $x$. This lower bound holds true with a probability of at least $1-\alpha$, and the estimation of the lower bound can be improved by increasing the number $n$ of purified samples that are classified. 

For $n$ classification trials, denote the number of runs where the prediction is correct as $n_c\le n$. Let $p:=n_c / n$ and denote $\operatorname{Beta}(\alpha ; n, p)$ as the $\alpha$-th quantile of a beta distribution with parameters $n$ and $p$. If we (safely) assume that

$p, n_c \sim Binomial\left(n, p\right)$, then the Clopper-Pearson estimation \cite{clopper_use_1934} allows us to say:
\begin{equation}
min(p_c(x))=\operatorname{Beta}\left(\alpha; n_c, n-n_c+1\right)
\end{equation}
with probability of at least $(1-\alpha)$. 

This estimation will be useful in establishing a starting point for certifiable robustness. Based on Corollary 1.1 in \cite{zeng_certified_2021}, rearranging equation (\ref{thm1}) tells us that 
\vspace{2mm}
\begin{equation}
    \mathbb{P}\big(g(x^\prime) = c\ |\  0.5 < min(p_c(x))-\beta \Delta \big) \ge 1-\alpha .
\end{equation}

Hence, if appropriate estimates for $\beta$ and $min(p_c(x))$ can be obtained, then we can determine the conditions under which our classifier is robust against any $d$-perturbed adversary, $x^\prime$; this robustness is guaranteed with a probably of at least $(1-\alpha)$; in this work, we set $\alpha = 0.05$ for calculating robustness certificates, so that we can compare to Zeng et al. \cite{zeng_certified_2021}.

\subsection{Empirical Evaluation of Robustness Certificates}

\begin{table}[h!]
\caption{Robustness certificates on AG News with different masking rates $m$.}
\centering

\begin{tabular}{|c|c|c|c||c|c|c|c|}
\hline
Method & Rate $m\%$ & Acc\% & MCB & Method & Rate $m\%$ & Acc\% & MCB\\
\hline 
\multirow{6}{*}{\shortstack[c]{RanMASK \\ \cite{zeng_certified_2021}}} 
& 0.4 & 96.2 & 1 &
    \multirow{6}{*}{\shortstack[c]{MaskPure 
    \\ (Ours)} } & 0.1 & 96.0 & 1 \\
& 0.5 & 95.7 & 1 &
    & 0.3 & 96.2 & 1 \\
& 0.6 & 95.7 & 2 &
    & 0.4 & 96.2 & 1 \\
& 0.7 & 94.5 & 2 &
    & 0.5 & 95.4 & 1\\
& 0.8 & 92.0 & 3 &
    & 0.6 & 94.0 & 2\\
& 0.9 & 91.1 & 5 &
    & 0.7 & 93.1 & 2\\ 
\hline
\end{tabular}
\label{table:agnews_robustness}
\end{table}


To verify the certified robustness of MaskPure for any given input, it is possible to follow a similar process to Zeng et al. \cite{zeng_certified_2021}, using Monte Carlo methods to estimate values of $p_c(x)$. Informed by this work, we similarly approximate $\beta \approx p_c(x)$ and run a series of trials to find $p_c(x)$. Results are shown in Table \ref{table:agnews_robustness}. There, MCB is ``Median Certified Robustness: the largest number, $d$, of perturbations that can be applied to samples in the dataset such that $>50\%$ of the texts are still classified correctly.

In Table \ref{table:agnews_robustness}, for Zeng et al. \cite{zeng_certified_2021}, maximum median certified robustness (MCB) was achieved when using $m=0.9$ or higher. For MaskPure, maximum MCB was achieved with a lower masking rate of 0.6, indicating the increased efficiency of MaskPure. Though the maximum MCB is a greater value for RanMASK, MaskPure obtains comparable MCB and accuracy when comparing at lower masking rates (i.e. RanMASK with $m = 0.4$ has approximately the same performance as MaskPure at $m=0.1$), which is better for computational cost. Note that this comparison is not entirely fair, since Zeng et al. \cite{zeng_certified_2021} used RoBERTA rather than BERT for their certificates.
  
\section{Conclusion}

Random purification has much potential for defending against adversarial inputs, as has been demonstrated in computer vision and by some pioneering works in NLP. By further exploring the effectiveness and benefits of stochastic purification in NLP, MaskPure contributes to filling this largely-unexplored gap in the literature. Our method demonstrates empirically-exceptional and certifiably-robust performance on the AG News and IMDB datasets when compared with previous defenses. This serves as a signal for future research to be conducted at the intersection of stochastic purification and improving robustness.

\section{Acknowledgements}
 This material is based upon work supported by the National Science Foundation under Grant No. 2050919. Any opinions, findings, and conclusions or recommendations expressed in this material are those of the authors and do not necessarily reflect the views of the National Science Foundation.

\bibliographystyle{splncs04}
\bibliography{Maskpure}

\begin{thebibliography}{10}
\providecommand{\url}[1]{\texttt{#1}}
\providecommand{\urlprefix}{URL }
\providecommand{\doi}[1]{https://doi.org/#1}

\bibitem{alshemali_improving_2020}
Alshemali, B., Kalita, J.: Improving the {Reliability} of {Deep} {Neural} {Networks} in {NLP}: {A} {Review}. Knowledge-Based Systems  \textbf{191},  105210 (Mar 2020)

\bibitem{belinkov_synthetic_2018}
Belinkov, Y., Bisk, Y.: {Synthetic} {and} {Natural} {Noise} {Both} {Break} {Neural} {Machine} {Translation}  (2018)

\bibitem{carlini_certified_2022}
Carlini, N., Tramer, F., Dvijotham, K.D., Rice, L., Sun, M., Kolter, J.Z.: ({Certified}!!) {Adversarial} {Robustness} for {Free}! (Sep 2022)

\bibitem{clopper_use_1934}
Clopper, C.J., Pearson, E.S.: The use of confidence or fiducial limits illustrated in the case of the binomial. Biometrika  \textbf{26}(4),  404--413 (1934)

\bibitem{cohen_certified_2019}
Cohen, J., Rosenfeld, E., Kolter, Z.: Certified {Adversarial} {Robustness} via {Randomized} {Smoothing}. In: Proceedings of the 36th {International} {Conference} on {Machine} {Learning}. pp. 1310--1320. PMLR (May 2019), iSSN: 2640-3498

\bibitem{croitoru_diffusion_2023}
Croitoru, F.A., Hondru, V., Ionescu, R.T., Shah, M.: Diffusion models in vision: A survey. IEEE Transactions on Pattern Analysis and Machine Intelligence  \textbf{PP},  1--20 (03 2023)

\bibitem{devlin_bert_2019}
Devlin, J., Chang, M.W., Lee, K., Toutanova, K.: {BERT}: {Pre}-training of {Deep} {Bidirectional} {Transformers} for {Language} {Understanding}. In: NAACL. pp. 4171--4186. Minneapolis, Minnesota (Jun 2019)

\bibitem{devvrit_voting_2020}
Devvrit, Cheng, M., Hsieh, C.J., Dhillon, I.: Voting based ensemble improves robustness of defensive models (Nov 2020), arXiv:2011.14031 [cs, stat]

\bibitem{ebrahimi_hotflip_2018}
Ebrahimi, J., Rao, A., Lowd, D., Dou, D.: {HotFlip}: {White}-{Box} {Adversarial} {Examples} for {Text} {Classification} (May 2018), arXiv:1712.06751 [cs]

\bibitem{gao_black-box_2018}
Gao, J., Lanchantin, J., Soffa, M.L., Qi, Y.: Black-{Box} {Generation} of {Adversarial} {Text} {Sequences} to {Evade} {Deep} {Learning} {Classifiers}. In: 2018 {IEEE} {Security} and {Privacy} {Workshops} ({SPW}). pp. 50--56 (May 2018)

\bibitem{goyal_survey_2023}
Goyal, S., Doddapaneni, S., Khapra, M.M., Ravindran, B.: A {Survey} of {Adversarial} {Defences} and {Robustness} in {NLP}. ACM Computing Surveys p. 3593042 (Apr 2023)

\bibitem{jia_certified_2019}
Jia, R., Raghunathan, A., Göksel, K., Liang, P.: Certified {Robustness} to {Adversarial} {Word} {Substitutions}. In: {EMNLP}-{IJCNLP}. pp. 4129--4142. Association for Computational Linguistics, Hong Kong, China (Nov 2019)

\bibitem{jiang_smart_2020}
Jiang, H., He, P., Chen, W., Liu, X., Gao, J., Zhao, T.: {SMART}: {Robust} and {Efficient} {Fine}-{Tuning} for {Pre}-trained {Natural} {Language} {Models} through {Principled} {Regularized} {Optimization}. In: ACL. pp. 2177--2190. Association for Computational Linguistics, Online (Jul 2020)

\bibitem{jin_is_2020}
Jin, D., Jin, Z., Zhou, J.T., Szolovits, P.: Is {BERT} {Really} {Robust}? {A} {Strong} {Baseline} for {Natural} {Language} {Attack} on {Text} {Classification} and {Entailment} (Apr 2020), arXiv:1907.11932 [cs]

\bibitem{jones_robust_2020}
Jones, E., Jia, R., Raghunathan, A., Liang, P.: Robust {Encodings}: {A} {Framework} for {Combating} {Adversarial} {Typos}. In: ACL. pp. 2752--2765. Association for Computational Linguistics, Online (Jul 2020)

\bibitem{kurakin_adversarial_2017}
Kurakin, A., Goodfellow, I., Bengio, S.: Adversarial {Machine} {Learning} at {Scale} (Feb 2017), arXiv:1611.01236 [cs, stat]

\bibitem{levine_robustness_2020}
Levine, A., Feizi, S.: Robustness {Certificates} for {Sparse} {Adversarial} {Attacks} by {Randomized} {Ablation}. AAAI  \textbf{34}(04),  4585--4593 (Apr 2020), number: 04

\bibitem{li_textbugger_2019}
Li, J., Ji, S., Du, T., Li, B., Wang, T.: {TextBugger}: {Generating} {Adversarial} {Text} {Against} {Real}-world {Applications}. In: {NDSS}, {San} {Diego}, {February}, 2019. The Internet Society (2019)

\bibitem{li_bert-attack_2020}
Li, L., Ma, R., Guo, Q., Xue, X., Qiu, X.: {BERT}-{ATTACK}: {Adversarial} {Attack} {Against} {BERT} {Using} {BERT} (Oct 2020), arXiv:2004.09984 [cs]

\bibitem{li_text_2023-1}
Li, L.N., Song, D.N., Qiu, X.N.: Text {Adversarial} {Purification} as {Defense} against {Adversarial} {Attacks}. In: Proceedings of the 61st {Annual} {Meeting} of the {Association} for {Computational} {Linguistics} (Jul 2023)

\bibitem{li_diffusion_2023}
Li, Y., Zhou, K., Zhao, W.X., Wen, J.R.: Diffusion {Models} for {Non}-autoregressive {Text} {Generation}: {A} {Survey} (May 2023), arXiv:2303.06574 [cs]

\bibitem{li_searching_2021}
Li, Z., Xu, J., Zeng, J., Li, L., Zheng, X., Zhang, Q., Chang, K.W., Hsieh, C.J.: Searching for an {Effective} {Defender}: {Benchmarking} {Defense} against {Adversarial} {Word} {Substitution}. In: EMNLP. pp. 3137--3147. Association for Computational Linguistics, Online and Punta Cana, Dominican Republic (Nov 2021)

\bibitem{maas_learning}
Maas, A.L., Daly, R.E., Pham, P.T., Huang, D., Ng, A.Y., Potts, C.: Learning word vectors for sentiment analysis. In: ACL-HLT. pp. 142--150. Association for Computational Linguistics, Portland, Oregon, USA (June 2011)

\bibitem{madry_towards_2019}
Madry, A., Makelov, A., Schmidt, L., Tsipras, D., Vladu, A.: Towards {Deep} {Learning} {Models} {Resistant} to {Adversarial} {Attacks} (Sep 2019), arXiv:1706.06083 [cs, stat]

\bibitem{miyato_adversarial_2021}
Miyato, T., Dai, A.M., Goodfellow, I.: Adversarial {Training} {Methods} for {Semi}-{Supervised} {Text} {Classification} (Nov 2021), arXiv:1605.07725 [cs, stat]

\bibitem{morris_textattack_2020}
Morris, J.X., Lifland, E., Yoo, J.Y., Grigsby, J., Jin, D., Qi, Y.: {TextAttack}: {A} {Framework} for {Adversarial} {Attacks}, {Data} {Augmentation}, and {Adversarial} {Training} in {NLP} (Oct 2020), arXiv:2005.05909 [cs]

\bibitem{nie_diffusion_2022}
Nie, W., Guo, B., Huang, Y., Xiao, C., Vahdat, A., Anandkumar, A.: Diffusion {Models} for {Adversarial} {Purification} (May 2022), arXiv:2205.07460 [cs]

\bibitem{sakaguchi_robsut_2017}
Sakaguchi, K., Duh, K., Post, M., Durme, B.V.: Robsut wrod reocginiton via semi-character recurrent neural network. In: {AAAI}. pp. 3281--3287. {AAAI}'17, San Francisco, California, USA (Feb 2017)

\bibitem{si_better_2021}
Si, C., Zhang, Z., Qi, F., Liu, Z., Wang, Y., Liu, Q., Sun, M.: Better {Robustness} by {More} {Coverage}: {Adversarial} and {Mixup} {Data} {Augmentation} for {Robust} {Finetuning}. In: Findings of {ACL}-{IJCNLP} 2021. pp. 1569--1576. Association for Computational Linguistics, Online (Aug 2021)

\bibitem{swenor_using_2021}
Swenor, A., Kalita, J.: Using {Random} {Perturbations} to {Mitigate} {Adversarial} {Attacks} on {Sentiment} {Analysis} {Models}. In: {International} {Conference} on {Natural} {Language} {Processing} ({ICON}). pp. 519--528. National Institute of Technology, Silchar, India (Dec 2021)

\bibitem{wolf_huggingfaces_2020}
Wolf, T., Debut, L., Sanh, V., Chaumond, J., Delangue, C., Moi, A., Cistac, P., Rault, T., Louf, R., Funtowicz, M., Davison, J., Shleifer, S., von Platen, P., Ma, C., Jernite, Y., Plu, J., Xu, C., Scao, T.L., Gugger, S., Drame, M., Lhoest, Q., Rush, A.M.: {HuggingFace}'s {Transformers}: {State}-of-the-art {Natural} {Language} {Processing} (Jul 2020), arXiv:1910.03771 [cs]

\bibitem{xiao_densepure_2023}
Xiao, C., Chen, Z., Jin, K., Wang, J., Nie, W., Liu, M., Anandkumar, A., Li, B., Song, D.: {DensePure}: {Understanding} {Diffusion} {Models} for {Adversarial} {Robustness} (Feb 2023)

\bibitem{ye_safer_2020}
Ye, M., Gong, C., Liu, Q.: {SAFER}: {A} {Structure}-free {Approach} for {Certified} {Robustness} to {Adversarial} {Word} {Substitutions}. In: ACL. pp. 3465--3475. Association for Computational Linguistics, Online (Jul 2020)

\bibitem{yoo_towards_2021}
Yoo, J.Y., Qi, Y.: Towards {Improving} {Adversarial} {Training} of {NLP} {Models} (Sep 2021), arXiv:2109.00544 [cs]

\bibitem{zeng_certified_2021}
Zeng, J., Xu, J., Zheng, X., Huang, X.: Certified robustness to text adversarial attacks by randomized [{MASK}]. Computational Linguistics  \textbf{49}(2),  395--427 (Jun 2023). \doi{10.1162/coli_a_00476}, \url{https://aclanthology.org/2023.cl-2.5}

\bibitem{zhang_character-level_2015}
Zhang, X., Zhao, J., LeCun, Y.: Character-level {Convolutional} {Networks} for {Text} {Classification}. In: Cortes, C., Lawrence, N., Lee, D., Sugiyama, M., Garnett, R. (eds.) NIPS. vol.~28. Curran Associates, Inc. (2015)

\bibitem{zhu_freelb_2020}
Zhu, C., Cheng, Y., Gan, Z., Sun, S., Goldstein, T., Liu, J.: {FreeLB}: {Enhanced} {Adversarial} {Training} for {Natural} {Language} {Understanding} (Apr 2020), arXiv:1909.11764 [cs]

\bibitem{zou_diffusion_2023}
Zou, H., Kim, Z.M., Kang, D.: Diffusion {Models} in {NLP}: {A} {Survey} (May 2023), arXiv:2305.14671 [cs]

\end{thebibliography}

\end{document}